\documentclass[letterpaper, 10 pt, conference]{ieeeconf}  

\IEEEoverridecommandlockouts                           
\usepackage{cite}
\usepackage{subcaption} 
\usepackage{amsmath,amssymb,amsfonts}
\usepackage{algorithmic}
\usepackage{graphicx}
\usepackage{textcomp}
\usepackage{booktabs}
\usepackage[table]{xcolor}
\usepackage{tabularx}
\usepackage{listings}
\usepackage{xcolor}
\usepackage{soul}   

\usepackage[colorlinks=true, linkcolor=blue]{hyperref} 

\usepackage{censor}
\usepackage{tcolorbox}
\tcbuselibrary{listings, breakable}

\definecolor{promptbg}{RGB}{248,248,248}
\definecolor{promptframe}{RGB}{180,180,180}

\newtcolorbox[auto counter]{promptbox}[2][]{%
    title=Template~\thetcbcounter: #2, 
    colback=promptbg,
    colframe=promptframe,
    boxrule=0.6pt,
    arc=3pt,
    left=6pt, right=6pt, top=6pt, bottom=6pt,
    breakable,
    #1 
}

\lstdefinestyle{promptstyle}{
    basicstyle=\ttfamily\small,
    breaklines=true,
    columns=fullflexible,
    keepspaces=true,
    showstringspaces=false,
    frame=none
}

\newcommand{\ours}{ELHPlan}
\def\BibTeX{{\rm B\kern-.05em{\sc i\kern-.025em b}\kern-.08em
    T\kern-.1667em\lower.7ex\hbox{E}\kern-.125emX}}
    
\title{\LARGE \bf
ELHPlan: Efficient Long-Horizon Task Planning for Multi-Agent Collaboration
}

\author{Shaobin Ling$^{1}$, Yun Wang$^{2}$, Chenyou Fan$^{3}$, Tin Lun Lam$^{1}$, and Junjie Hu$^{1}$*
	\thanks{*Corresponding author: Junjie Hu}
	\thanks{$^{1}$Shaobin Ling, Tin Lun Lam, and Junjie Hu are with The Chinese University of Hong Kong, Shenzhen, China. 
		{\tt\small ab.pixel.pixel@gmail.com, tllam@cuhk.edu.cn, hujunjie@cuhk.edu.cn}}%
	\thanks{$^{2}$Yun Wang is with City University of Hong Kong, Hong Kong, China.
		{\tt\small wangy978@mail2.sysu.edu.cn}}%
	\thanks{$^{3}$Chenyou Fan is with South China Normal University, Guangzhou, China.
		{\tt\small fanchenyou@scnu.edu.cn}}%
}

\renewcommand{\hl}[1]{#1}

\begin{document}

\maketitle
\thispagestyle{empty}
\pagestyle{empty}

\begin{abstract}
Large Language Models (LLMs) enable intelligent multi-robot collaboration but face fundamental trade-offs: 
open-loop methods that compile tasks into formal representations for external executors produce sound plans but lack adaptability in partially observable environments, while iterative methods incur prohibitive computational costs that scale poorly with team size and task complexity. In this paper, we propose 
\textbf{E}fficient \textbf{L}ong-\textbf{H}orizon \textbf{Plan}ning  
(\ours), a novel framework that introduces Action Chains, sequences of actions explicitly bound to sub-goal intentions, as the fundamental planning primitive. ELHPlan operates via a cyclical process: 1) constructing intention-bound action sequences, 2) proactively validating for conflicts and feasibility, 3) refining issues through targeted mechanisms, and 4) executing validated actions. This design balances adaptability and efficiency by providing 
intention-bound action sequences with longer lookahead 
while avoiding expensive full re-planning. 
We further advocate comprehensive efficiency metrics, including token consumption and planning time, to more holistically evaluate multi-agent collaboration.
Our experiments on benchmarks TDW-MAT and C-WAH demonstrate that ELHPlan achieves comparable task success rates while consuming only \hl{30\%--40\%} of the tokens required by state-of-the-art methods. 
Our research establishes a new efficiency-effectiveness frontier for LLM-based multi-agent planning systems.

\end{abstract}

\section{Introduction}
\label{sec:introduction}

Coordinating multiple agents to collaboratively accomplish complex tasks in dynamic environments represents a fundamental challenge in modern robotics, requiring sophisticated planning algorithms, effective communication protocols, and robust coordination mechanisms. 
Recent advances in Large Language Models (LLMs) have marked a significant step towards intelligent robotics, endowing robots with the ability to understand natural language instructions and reason about complex action sequences in collaborative environments. 

\begin{figure}[t]
    \centering
    \includegraphics[width=0.4\textwidth]{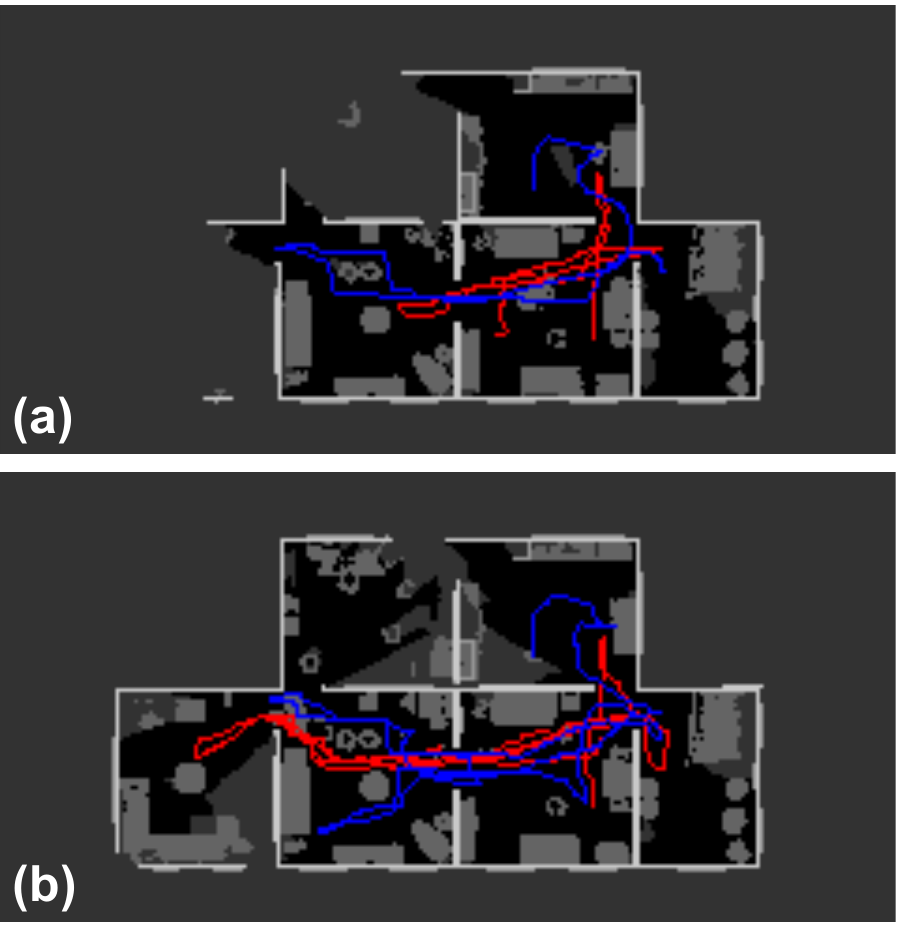}
    \caption{
    Trajectories and explored maps of
    CoELA~\cite{zhang2024building} (a) and our approach (b) in the same task. The blue trajectory represents agent 1's path. The red trajectory represents agent 2's path. 
    The trajectories illustrate less redundant overlap under comparable token budgets (22,882 vs. 23,042 tokens), highlighting improved coordination efficiency.
    }
    \label{fig:trajectories_comparison}
\end{figure}


Existing LLM-based planning methods face a fundamental trade-off between plan quality and environmental adaptability. 
Open-loop methods~\cite{zhao2023large, liu2025delta, singh2024twostep} compile tasks into formal representations for external executors, producing sound plans but assuming complete world models at planning time. 
This limits their applicability to partially observable environments. Conversely, iterative methods~\cite{yao2023react, prasad-etal-2024-adapt, zhou2024llm} generate actions step-by-step through frequent LLM queries, enabling adaptation to dynamic settings but incurring prohibitive computational overhead. 
\hl{The challenge is further compounded in multi-agent scenarios, where coordination effectiveness depends on mutual understanding of action intentions.
 Such intentions function as critical coordination signals, enabling agents to avoid resource conflicts and task duplication.
Existing iterative approaches typically rely on either explicit communication of intention through natural language dialogues or implicit intention inference among multiple agents.}
However, explicit communication through natural language dialogues~\cite{mandi2024roco, li2023camel, shi2024opex} incurs substantial token costs, while implicit intention inference~\cite{zhang2024proagent, Liu2024LLM-Powered} often leads to coordination failures \hl{due to intention misinterpretation}. 
These limitations are particularly pronounced in long-horizon tasks\footnote{
In this paper, we use long-horizon to denote tasks with a benchmark-specified maximum step budget \(H\) (e.g., \(H \ge 150\) ) and multiple sub-goals, where success depends on extended multi-step decision making. } and motivate a fundamental question: \textit{How can we achieve the flexibility of iterative planning in multi-agent scenarios while minimizing token consumption?} 

\begin{figure}[t]
    \centering
    \includegraphics[width=0.4\textwidth]{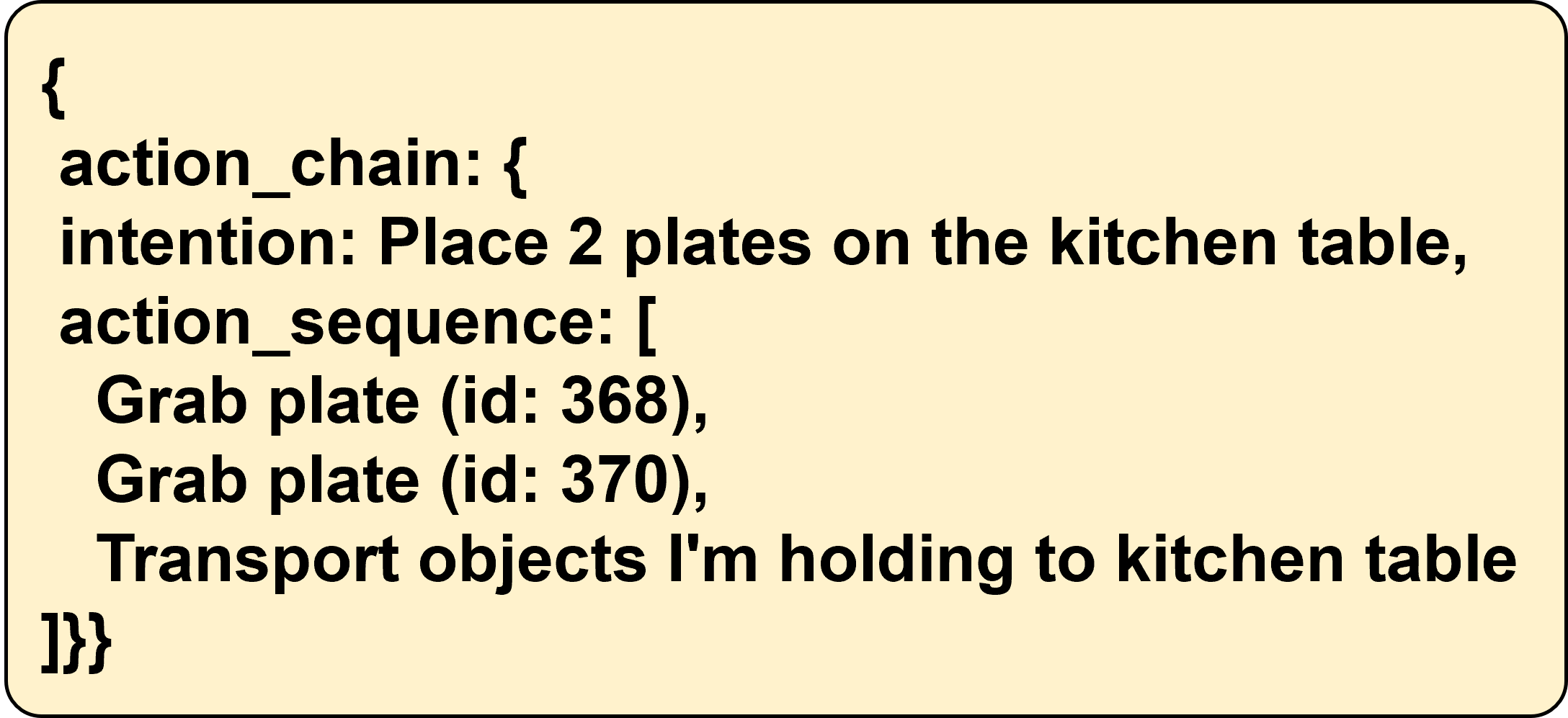}
    \caption{An example of an Action Chain presented in JSON-like format, containing the action sequence and  intentions.
    }
    \label{fig:action_chain}
\end{figure}

To address this challenge, we propose \textit{Action Chain}, a minimal planning unit that represents multi-step action sequences together with explicit sub-goal intentions, as illustrated in Figure~\ref{fig:action_chain}. 
\hl{Unlike prior approaches that treat action planning and intention communication as separate processes, our Action Chain unifies them by explicitly binding each action sequence to a declared sub-goal intention.
}
We argue that binding explicit sub-goal intentions to action sequences can serve as a lightweight coordination mechanism: rather than requiring agents to infer collaborators' goals through separate reasoning modules~\cite{zhang2024proagent, Liu2024LLM-Powered} or costly natural language dialogues ~\cite{mandi2024roco,li2023camel,shi2024opex}, each agent can directly read the declared intentions embedded in its partners' Action Chains. 
This design eliminates the need for additional intention-inference queries to the LLM, thereby reducing both token consumption and inference latency. 
\hl{We validate this hypothesis empirically through ablation studies in task efficiency. }(Section~\ref{sec:ablation_study}). 

While Action Chains offer these advantages, their practical deployment necessitates robust construction and refinement processes. 
Therefore, we propose \textbf{E}fficient \textbf{L}ong-\textbf{H}orizon Task \textbf{Plan}ning (ELHPlan), 
a multi-agent planning framework designed to construct and refine Action Chains from minimal planning units into complete robotic planning solutions. 
ELHPlan operates through a four-stage cyclical process: First, the planning module generates Action Chains for each agent. Second, the validation module detects infeasible actions and conflicts. Third, if issues are identified, the refinement stage creates new or extends existing Action Chains. Finally, validated actions proceed to execution. The framework iterates until all tasks are completed. 
Beyond task-level metrics such as Transport Rate and Move Distance, we systematically measure Inference Time and Token Consumption across all methods\cite{seo2024reveca,zhang2024building}, enabling a direct comparison of the efficiency-effectiveness trade-off critical to practical deployment. 
To validate our approach, we conduct comprehensive experiments on two embodied environments: ThreeDWorld Multi-Agent Transport (TDW-MAT)  and Communicative Watch-And-Help (C-WAH). Our experimental results demonstrate that our method achieves comparable performance to state-of-the-art approaches while improving computational efficiency in long-horizon multi-agent tasks.


The main contributions of this paper are:
\begin{itemize}

    \item  \textbf{Action Chain Representation:} We introduce Action Chains as a novel planning primitive that couples multi-step action sequences with explicit sub-goal intentions. \hl{This representation enables agents to commit to extended strategies under partial observability while exposing their intentions to collaborators without additional inference queries.}

    \item  \textbf{Proactive Validation-Refinement:} We design a comprehensive refinement mechanism that proactively validates Action Chains for feasibility and inter-agent conflicts before execution, thereby supporting flexible iterative planning with efficient intention sharing and reduced token consumption.

    \item Our method achieves competitive performance across multiple benchmarks, while reducing token cost to only 30\%--40\% of that of state-of-the-art methods.
    

\end{itemize}
    

\section{Related Work} 

\subsection{LLMs for Task Planning} 

Recent studies have demonstrated the effectiveness of LLMs in robotic task planning. 
These approaches can be classified based on their adaptability to dynamic environments and computational efficiency.

\textbf{Open-loop Planning.} 
A prominent line of work uses LLMs to generate a complete plan covering the entire task before execution begins, without subsequent adaptation to environmental feedback. These methods differ in their target formalism: some translate tasks into symbolic specifications such as PDDL for classical planners.  
LLM+P~\cite{zhao2023large} translates natural language problem descriptions into PDDL problems, and DELTA~\cite{liu2025delta} uses scene graphs as an environment representation and formal planning to decompose long-term tasks. TwoStep~\cite{singh2024twostep} adopts the paradigm to multi-agent settings. 
\hl{Others generate executable code for program interpreters.} 
Code as Policies~\cite{liang2023code} and ProgPrompt~\cite{singh2023progprompt} generate Python code that invokes robot primitives through an interpreter,
and Voxposer~\cite{huang2023voxposer} synthesizes 3D value maps via code for manipulation tasks. 
While these approaches benefit from the completeness and soundness guarantees of classical planners, they fundamentally assume access to complete and accurate world models at planning time, limiting their applicability to partially observable and dynamic  environments.  
In contrast, our approach operates under partial observability and constructs plans based on accumulated observations. 


\textbf{Iterative Planning.} 
While open-loop methods prioritize plan soundness at the cost of environmental adaptability, iterative methods take the opposite trade-off 
\hl{, generating actions step-by-step through frequent LLM queries}~\cite{prasad-etal-2024-adapt,zhou2024llm}.
\hl{SayCan{~\cite{ichter2022do}} grounds LLM outputs by scoring each candidate skill against pre-trained affordance functions at every decision step. However, this per-step ranking is inherently single-step, producing no multi-step action commitment. It also operates in a single-agent setting, offering no mechanism to expose planning intentions to collaborating agents.}
\hl{A representative example is LLM-Planner{~\cite{song2023llmplanner}}, which dynamically adjusts plans based on environmental feedback. However, its re-planning is reactive and monolithic, discarding the entire remaining plan upon failure. It also operates solely in single-agent settings without intention-level coordination. } 
\hl{To reduce redundant reasoning, knowledge-augmented approaches build reusable planning knowledge from past experience. }
Retrieval-augmented methods~\cite{xie2024embodied, Kagaya2024RAPRP, xu2024prag} maintain non-parametric experience memories and condition plan generation on retrieved relevant episodes, improving sample efficiency without model retraining. 
However, these methods typically require substantial pre-collected data or extended exploration. 
Our approach bridges open-loop planning and iterative planning paradigms: Action Chains provide structured multi-step formal commitments while remaining incrementally refinable under partial observability. 

\subsection{Multi-agent Cooperation via LLMs} 
Recent studies have increasingly explored the multi-agent cooperation via LLMs~\cite{Liu2024LLM-Powered,li2025large, Chen2024Scalable}, which can be broadly classified into  decentralized  or centralized cooperation.

\textbf{Decentralized cooperation} relies on natural language communication among agents. 
Role-playing dialogue systems such as Roco~\cite{mandi2024roco}, CAMEL~\cite{li2023camel}, and OPEX~\cite{shi2024opex} enable agents to negotiate task allocation through natural language exchanges. 
Beyond role-playing dialogue, CoELA~\cite{zhang2024building} introduces a modular architecture with explicit communication and cooperative planning modules for long-horizon household tasks. 
REVECA~\cite{seo2024reveca} further improves coordination through adaptive planning with trajectory-based validation in similar multi-objective settings. 
Both specifically target long-horizon, multi-objective household tasks, making them representative baselines for our evaluation setting. 
Another line of work investigates intention inference, where agents reason about or predict collaborators' intentions and then take that into consideration for planning~\cite{zhang2024proagent, Liu2024LLM-Powered}.
These methods provide rich semantics but require frequent LLM queries for communication or inference, leading to inefficiency in long-horizon tasks.

\textbf{Centralized cooperation} introduces an LLM-based coordinator for task allocation and monitoring. 
Smart-LLM~\cite{kannan2024smart} integrates task decomposition and coalition formation. 
LLaMAR~\cite{nayak2024longhorizon} adopts a modular architecture with correction and verification. 
Shared memory mechanisms, such as REBEL~\cite{gupte2024rebel}, enable agents to reuse prior experiences for improved task allocation. 
Another line of research addresses task dependencies explicitly, using directed acyclic graphs to decompose high-level instructions into well-coordinated subtasks for multi-robot execution~\cite{wang2024dart, obata2024lip}.

Decentralized schemes are flexible but costly, while centralized schemes are efficient but less adaptive. 
Our work employs centralized Action Chain construction for efficient planning, while shared memory propagates each agent's intentions and observations to enable decentralized adaptation without costly inter-agent dialogue. 
Unlike implicit intention inference methods~\cite{zhang2024proagent, Liu2024LLM-Powered}, ELHPlan embeds explicit intentions directly within Action Chains, avoiding separate inference queries and reducing coordination overhead.  

\begin{figure*}
\centering
\includegraphics[width=\textwidth]{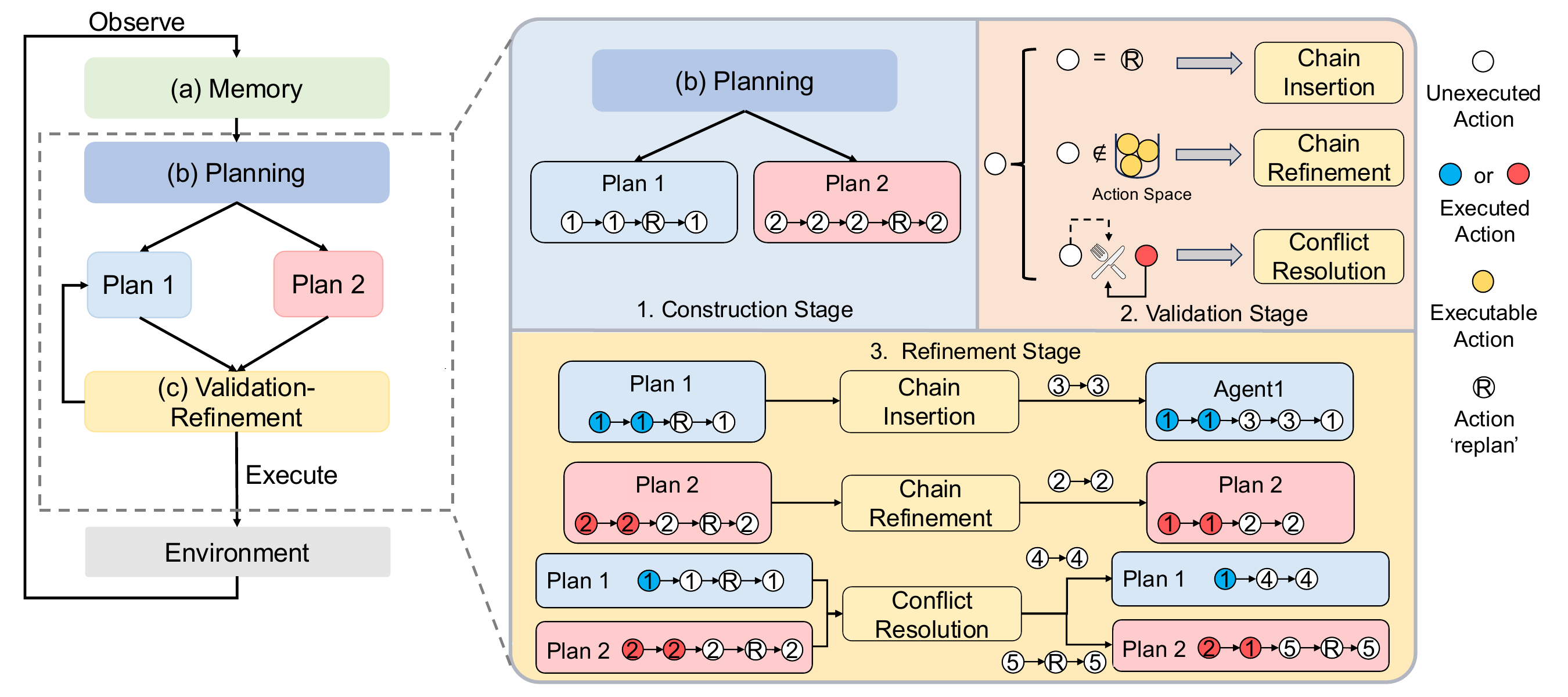}
\caption{The framework consists of a three-stage workflow (Memory, Planning, and Validation-Refinement) and corresponding processing flows. 
\hl{Each numbered node represents an action, and the number indicates the planning iteration that generated it: nodes labeled 1 were produced by the first planning call, nodes labeled 2 by the second, and so on. Nodes sharing the same number collectively form one Action Chain. }
In the refinement stage, we assume that agents' plans encounter different cases of refinement at varying stages of action execution progress, thereby demonstrating that distinct refinement methods exert differential impacts on the original plan. } 
\label{fig2:pipeline_a}
\end{figure*} 

\section{Approach}

The multi-agent collaboration can be formulated as a process of Partially Observable Markov Decision (POMDP), which captures sequential decision-making under uncertainty. The environment is described by a state space $S$, an action space $A$ and an observation space $O$. Actions follow a transition function $T$, which governs how the state evolves. 
\hl{An event-driven centralized planner collects the latest observations from agents upon planning requests and maintains a shared memory of states to guide decision making. }
The system proceeds in discrete steps until all sub-goals are achieved or the step limit of $H$ is reached. 

\subsection{System Overview}
ELHPlan strategically decomposes complex tasks into coordinated action chains, as illustrated in Figure~\ref{fig2:pipeline_a}, where the collaborative behavior of two agents is depicted.



(a) Memory Module receives structured partial observations $o_t^i \in \mathcal{O}$ about environmental data (objects, states, agent status) and maintains persistent storage of task instructions, observation histories $\{o^i_1, \dots, o^i_t\}$, and \hl{shared state estimate $\hat{s}_t$ aggregated across agents.} 
It outputs contextualized prompts by implementing rule-based retrieval that selects relevant few-shot examples from a predefined case library, providing domain-specific guidance for planning. 
(b) Planning Module takes contextualized prompts and generates agent-specific Action Chains. Each action chain 
is bound to an explicit sub-goal intention 
, with strategic `replan' placeholders inserted to handle environmental uncertainty. 
(c) Validation-Refinement Module receives unexecuted action from each agents' Action Chains and \hl{current state estimate $\hat{s}_t$}, performing feasibility checks and conflict detection. It outputs validated executable action. 
The Observation and Execution Modules are adopted from prior frameworks~\cite{zhang2024building}. 

\subsection{Planning with Action Chain}

The Action Chain planning process operates through three distinct stages, each designed to maximize efficiency while ensuring robust multi-agent coordination:

The central concept in ELHPlan is the Action Chain, a sequence of actions with shared intention to solve sub-goal. The chain length adapts dynamically to sub-goal's complexity. All modules are organized around the Action Chain, which governs how Action Chains are constructed, validated, and refined. 

\textbf{Stage 1: Construction.} 
The Planning Module generates Action Chains via single LLM calls, producing coherent action sequences bound by explicit intentions. The LLM receives structured prompts containing the current goal, shared memory context, and recent observations. 
Key features include dynamic chain length adaptation based on sub-goal complexity and strategic insertion of `replan' placeholders where environmental uncertainty requires deferred detailed planning. 
\hl{In Fig. {~\ref{fig2:pipeline_a}}, each Action Chain is annotated with an index corresponding to the planning iteration that produced it. For example, all actions labeled 1 originate from the first planning call, while actions labeled 3 are generated by the third call. This indexing makes the temporal provenance of each action explicit and visually traceable across the three stages. }
When an agent's current Action Chain is exhausted, the system re-enters the construction stage to generate new Action Chains that extend the agent's plan.

\textbf{Stage 2: Validation.} 
First, the latest unexecuted action from the Action Chain is matched against the action `replan'. If it corresponds to `replan', the system initiates Chain Insertion. Otherwise, the action undergoes a feasibility assessment with respect to the current world states. 
The system maintains a dynamic action space $A(\hat{s}_t) \subseteq A$, encompassing all executable operations given the current state estimate $\hat{s}_t$, thereby ensuring that only valid actions proceed to execution. 
Actions are validated by checking whether their preconditions are satisfied 
under $\hat{s}_t$ (e.g., $\mathrm{grasp}(\mathrm{apple})$ requires that the apple's spatial coordinates are known in $\hat{s}_t$). 
Should an action prove infeasible, the system triggers Chain Refinement. 
Finally, the framework detects inter-agent conflicts when multiple agents select actions $a_t^{(i)}, a_t^{(j)} \in A$ that target identical objects, as such concurrent actions violate the single-occupancy constraint implicit in the transition function $T$. 
Upon detection of inter-agent conflicts, the system invokes Conflict Resolution.

\textbf{Stage 3: Refinement}
When validation identifies issues, three specialized mechanisms address them. 
Chain Refinement removes or replaces inefficient actions while preserving the original chain intention. 
Conflict Resolution reconstructs and modifies the Action Chains of two agents in conflict to prevent sub-goal overlaps.
Chain Insertion creates new Action Chains to replace action `replan' using updated environmental observations. 


This three-stage cycle creates a compositional planning structure that adapts dynamically to environmental changes while maintaining agent coordination. 
Detailed prompt templates are provided in the supplementary file (Templates 1-4).

\section{Experiment}

\subsection{Experiment Setup}
\label{sec:setup}
Following previous methods~\cite{seo2024reveca,zhang2024building}, we evaluate ELHPlan on two indoor simulation environments designed for household tasks requiring long-horizon planning: TDW-MAT and C-WAH. 
Both environments share several key characteristics: they are partially observable and initially completely unknown to the agents, requiring agents to handle dynamic environmental changes and coordination challenges. 
\hl{Consistent with prior work, the simulators provide ground-truth world states, such as object locations, agent positions, as observations to all methods equally.} 
Both experiment settings satisfy our long-horizon definition: 
TDW-MAT operates with $H = 3{,}000$ steps and 10 sub-goals per task, while C-WAH uses $H = 150$ steps with 3–5 sub-goals, both requiring agents to sequence and coordinate multiple sub-goals within a fixed planning horizon. 

\begin{figure*}
    \centering
    \includegraphics[width=\textwidth]{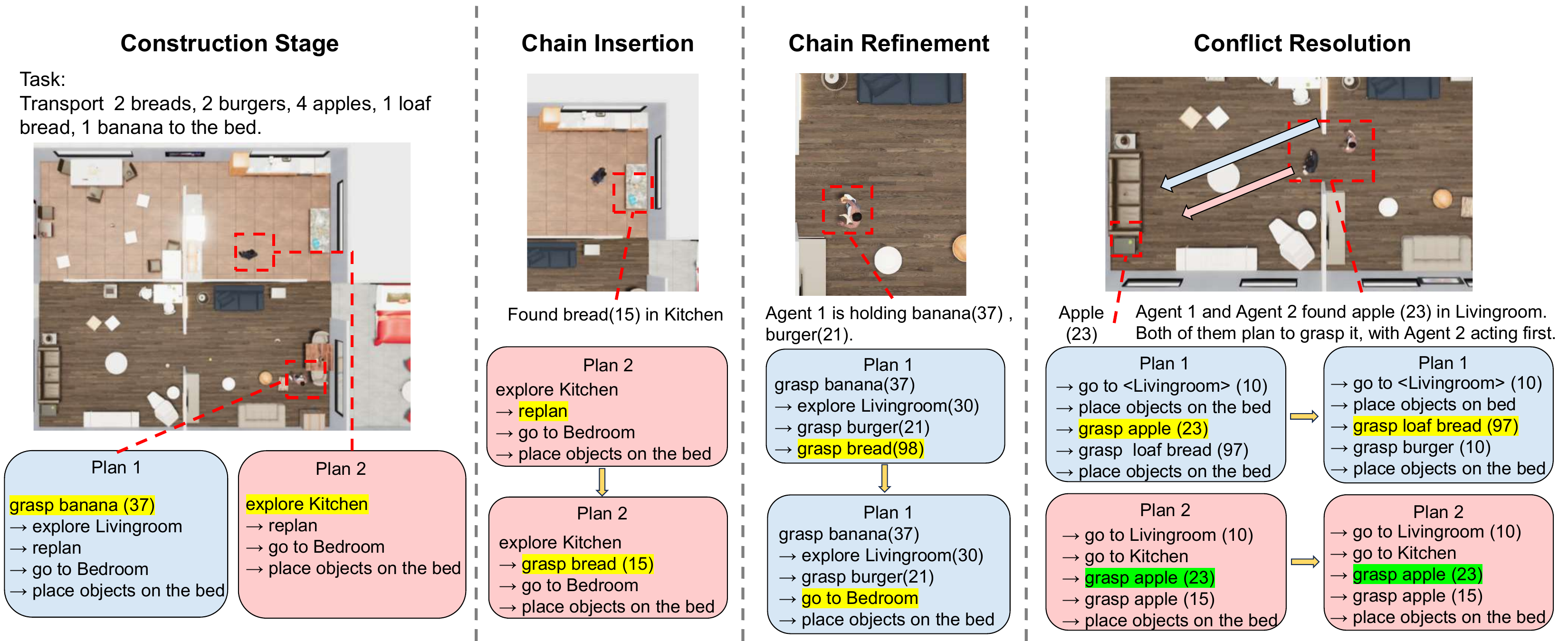}
    \caption{
Illustrative example of the ELHPlan during the Construction and Validation–Refinement stages. 
Initial task allocations are generated and assigned to two agents. 
Agent 2 performs a Chain Insertion triggered by action `replan', while Agent 1 subsequently refines its action chain to address an infeasible action. Finally, the conflict arising from both agents attempting to `grasp apple (23)' is resolved. 
The yellow highlight denotes the action scheduled for execution, whereas the green highlight denotes the action currently in execution. 
Identical items and room names are distinguished by numerical labels.
}
    \label{fig:fig3}
\end{figure*}

We conduct experiments on 24 test tasks in TDW-MAT and 10 test tasks in C-WAH to ensure comprehensive performance assessment. 
To evaluate multi-agent scalability, we test with team sizes of $N \in \{2, 3, 4, 5\}$ agents. 
Three backbone LLMs are considered: GPT-4o{~\cite{openai2024gpt4o}}, GPT-4o-mini{~\cite{openai2024gpt4o}}, and Llama~3.1{~\cite{grattafiori2024llama3}}. 
Cloud-based models are accessed via the OpenAI API.  Llama~3.1 runs locally on a workstation with an Intel Core i9-12900K CPU, 64 GB RAM, and dual NVIDIA RTX~3090,Ti GPUs. 
Each configuration is run for 5 independent trials. 
Before presenting the quantitative results, we first provide an example in Figure~\ref{fig:fig3} to demonstrate how the algorithm behaves during the Construction, Validation-Refinement stages. 

\subsection{Metric}
We evaluate agent performance using a comprehensive set of metrics that assess both computational efficiency and planning quality across the two simulation environments. Our evaluation framework employs both common metrics applicable across environments and specialized metrics tailored to each environment's specific objectives. 


Existing evaluation frameworks for multi-agent systems typically rely on task-specific performance indicators. 
\textbf{Transport Rate (TR)} is used in TDW-MAT to measure the percentage of items successfully transported to target locations within the step limit, indicating task completion efficiency. 
\textbf{Move Distance (MD)} calculates the average movement distance (in meters) per task, providing a measure of agents' plan quality. 
\textbf{Simulation Steps (SS)} are employed in C-WAH to count the total environment steps required to complete all sub-goals, with lower values indicating better efficiency in achieving overall objectives within the step constraints.    
While these conventional metrics provide valuable insights into task outcomes, they fail to capture computational costs that critically determine practical deployability. 
For LLM-based multi-agent systems, inference latency and token expenditure directly impact real-time responsiveness and operational cost—dimensions that have not been systematically benchmarked in prior evaluations on TDW-MAT and C-WAH. 
To address this gap, we incorporate two metrics. 
\textbf{Inference Time (IT)} measures the total computational time (in seconds) required for generating action plans, reflecting the computational efficiency of the reasoning process. 
\textbf{Token Consumption (TC)} tracks the cumulative number of input and output tokens consumed by LLM calls during task execution, quantifying the linguistic resource cost. 



\subsection{Overall Results}

\begin{figure*}[htbp] 
    \centering
    \includegraphics[width=\textwidth]{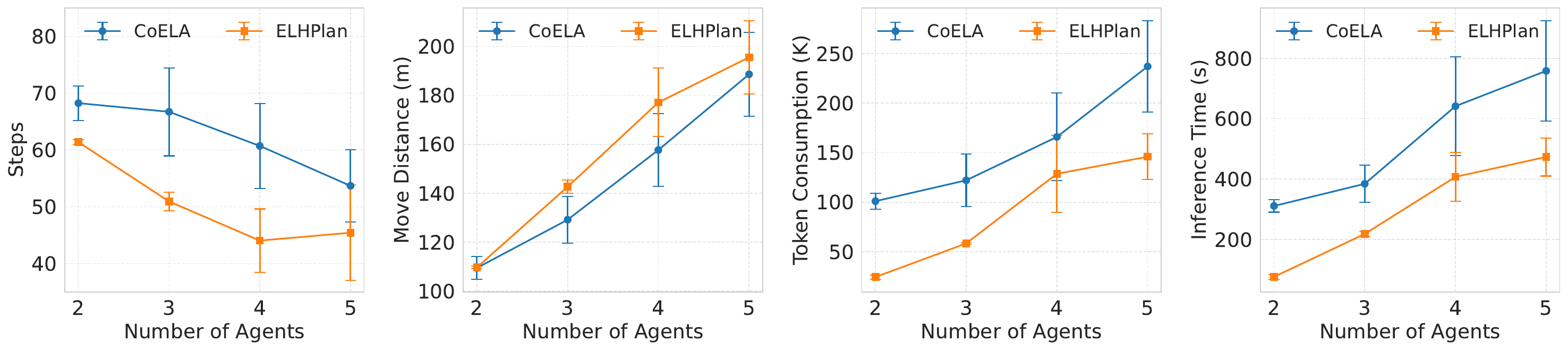} 
    \caption{Performance when varying the number of robots using GPT‑4o as the LLM backbone.}
    \label{fig:mr_result}
\end{figure*}

\begin{table*}[ht]
    \centering
    \setlength{\tabcolsep}{12pt} 
    \caption{Comparative Experimental Results in TDW-MAT Environment \hl{with 2 agents}. The best and second-best results for each metric are highlighted in \textbf{bold} and \underline{underlined}, respectively.} 
    \begin{tabular}{@{}lccccc@{}}
        \toprule
        \textbf{Method} & \textbf{LLMs} & \textbf{TR$\uparrow$} & \textbf{MD$\downarrow$} & \textbf{TC (K)$\downarrow$} & \textbf{IT (s)$\downarrow$} \\ \midrule
        CoELA   & Llama3.1    & 55.58 $\pm$ 4.55 & 307.28 $\pm$ 2.23 & 94.12 $\pm$ 1.18 & 496.66 $\pm$ 3.86 \\
        CoELA   & GPT-4o    & 71.72 $\pm$ 3.64 & \textbf{275.88 $\pm$ 5.03} & 76.28 $\pm$ 0.55 & 257.48 $\pm$ 2.85 \\
        CoELA   & 4o-mini    & 77.50 $\pm$ 4.12 & \underline{278.65 $\pm$ 4.04} & 70.18 $\pm$ 3.01 & 330.56 $\pm$ 68.73 \\
        \midrule
        REVECA  & 4o-mini  & \textbf{87.00} & /    & /        & /       \\ \midrule
        \ours   & Llama3.1    & 66.25 $\pm$ 1.77 & 306.58 $\pm$ 5.60 & 44.50 $\pm$ 3.55 & 372.07 $\pm$ 5.11 \\
        \ours   & GPT-4o    & \underline{79.17 $\pm$ 1.06} & 295.65 $\pm$ 3.01 & \underline{32.92 $\pm$ 0.26} & \underline{237.37 $\pm$ 0.48} \\        
        \ours   & 4o-mini    & 78.54 $\pm$ 5.60 & 300.57 $\pm$ 0.19 & \textbf{29.64 $\pm$ 1.04} & \textbf{209.14 $\pm$ 7.65} \\
    \bottomrule
    \end{tabular}
    \label{tab:tdwmat_combined_results}
\end{table*}

\begin{table*}[ht]
    \centering
    \setlength{\tabcolsep}{12pt} 
    \caption{Comparative Experimental Results in C-WAH Environment \hl{with 2 agents}. The best and second-best results for each metric are highlighted in \textbf{bold} and \underline{underlined}, respectively.}
    \footnotesize 
    \setlength{\tabcolsep}{2pt} 
    \begin{tabular}{@{}lccccc@{}}
        \toprule
        \textbf{Method} & \textbf{LLMs} & \textbf{SS$\downarrow$} & \textbf{MD$\downarrow$} & \textbf{TC (K)$\downarrow$} & \textbf{IT (s)$\downarrow$} \\ \midrule
        CoELA   & Llama3.1  & 80.15 $\pm$ 3.61 & 135.44 $\pm$ 6.49 & 119.46 $\pm$ 7.25 & 577.84 $\pm$ 30.00 \\
        CoELA   & GPT-4o    & 55.50 $\pm$ 0.42 & 97.79 $\pm$ 0.81 & 71.69 $\pm$ 0.19 & 197.61 $\pm$ 30.33 \\
        CoELA   & 4o-mini   & 68.25 $\pm$ 3.05 & 109.49 $\pm$ 4.60 & 101.00 $\pm$ 8.15 & 310.97 $\pm$ 20.50 \\ 
        \midrule
        REVECA  & Llama 3.1 & 59.56 $\pm$ 4.61   & \underline{79.10 $\pm$ 3.54}   & 158.44  $\pm$ 9.01  & 369.48 $\pm$  31.36 \\
        REVECA & GPT-4o & \textbf{49.68 $\pm$ 1.07} & 86.4 $\pm$ 0.89 & 72.51 $\pm$ 2.52 & 411.25 $\pm$ 8.27 \\ 
        REVECA & 4o-mini & 55.31 $\pm$ 2.08 & \textbf{76.42 $\pm$ 1.48} & 79.27 $\pm$ 2.52 & 374.53 $\pm$ 4.56 \\       \midrule
        \ours & Llama 3.1 & 73.47 $\pm$ 1.10 & 130.54 $\pm$16.36 & 32.86 $\pm$ 2.03  & 352.81 $\pm$ 23.00 \\
        \ours & GPT-4o & \underline{54.10 \(\pm\) 3.60} & 108.98 \(\pm\) 1.32 & \textbf{22.22 \(\pm\) 4.09} & \textbf{ 78.62 \(\pm\) 6.83} \\
        \ours & 4o-mini  & 58.58 \(\pm\) 2.42 & 111.26 \(\pm\) 5.82 & \underline{22.39 \(\pm\) 4.54} & \underline{84.98 \(\pm\) 20.86} \\
        \bottomrule
    \end{tabular}
    \label{tab:cwah_results}
\end{table*}

We compare ELHPlan against two state-of-the-art LLM-based multi-agent planning approaches:

CoELA~\cite{zhang2024building}is a collaborative planning framework that uses explicit communication and action coordination between agents. 
REVECA~\cite{seo2024reveca} is a recent method focusing on multi-agent coordination through structured reasoning and verification. 
All methods are evaluated with multiple LLMs to ensure fair comparison across different model capabilities. 
Due to the unavailability of REVECA's source code for TDW-MAT, only the TR metric is reported for this method.

\textbf{ELHPlan considerably outperforms the baseline methods in terms of computational resource consumption and response time.}
As shown in Table~\ref{tab:tdwmat_combined_results}, within the TDW-MAT environment, ELHPlan combined with GPT-4o-mini achieved the lowest token consumption (29.64K) and inference time (209.14s), reducing them by 57.7\% and 36.7\%, respectively, compared with the best configuration of CoELA (GPT-4o-mini with TR=77.5\%). 
In the C-WAH environment (Table~\ref{tab:cwah_results}), ELHPlan similarly maintained an advantage in computational efficiency, with GPT-4o achieving the lowest token consumption (22.22K) and inference time (78.62s), representing reductions of 69.3\% and 80.9\% compared with the best performance of REVECA using GPT-4o. 

\textbf{ELHPlan achieves the optimal balance between computational efficiency and task performance.} Although it slightly underperforms REVECA in terms of pure task success rate, its substantial advantage in resource consumption renders ELHPlan far more practical for real-world applications. In particular, in multi-agent collaboration scenarios requiring real-time responsiveness, ELHPlan’s low-latency property (with inference time of 78.62s compared to REVECA's 411.25s, approximately one-fifth) provides significant practical value. 
Moreover, ELHPlan demonstrates remarkable token efficiency, consuming  30.6\% of the tokens required by REVECA (22.22K vs 72.51K tokens in C-WAH environment) while maintaining comparable performance.

The performance of different methods varies significantly with the underlying LLM choice, revealing limitations of current LLM-based planning systems. 
\textbf{ELHPlan demonstrates consistency across LLM configurations}. 
For example, switching from Llama3.1 to GPT-4o improves the TR by 12.92\% (from 66.25\% to 79.17\%) in the TDW-MAT environment, while reducing token consumption from 44.50K to 32.92K, yielding a 26.02\% efficiency gain. 
In contrast, CoELA exhibits much higher sensitivity, with transport rates fluctuating by 21.92\% (55.58-77.50) across LLMs. 
This stability is consistently observed across environments, with ELHPlan showing substantially lower variance in key metrics. 
Such cross-LLM stability has practical implications, reducing dependence on specific LLMs and ensuring reliable performance under model switching.

Although ELHPlan excels in efficiency and robustness, its navigation results reveal suboptimal spatial reasoning. 
In both TDW-MAT and C-WAH, it generates longer paths than the best-performing methods, likely due to insufficient fine-grained optimization in its hierarchical planning and a stronger tendency toward exploratory actions. 

\textbf{ELHPlan demonstrates superior scalability with increasing team size.} 
As shown in Fig.{~\ref{fig:mr_result}}, we evaluate both methods with GPT-4o on C-WAH as the number of agents scales from 2 to 5. ELHPlan achieves comparable or lower Simulation Steps across all team sizes, confirming that additional agents are effectively utilized. More importantly, ELHPlan maintains a near-constant resource footprint: 
\hl{ its Token Consumption remains below 146K and Inference Time stays around 473s even at $N=5$, 
whereas CoELA's TC and IT grow sharply to over 237K and 764s, respectively. This advantage stems from the Action Chain mechanism, where each agent commits to multi-step plans per LLM call, making the marginal cost of adding agents negligible.
However, Move Distance increases for both methods with more agents, with ELHPlan exhibiting higher MD, consistent with the spatial reasoning limitation discussed above. }

\begin{figure}[t]
    \centering
    \includegraphics[width=0.5\textwidth]{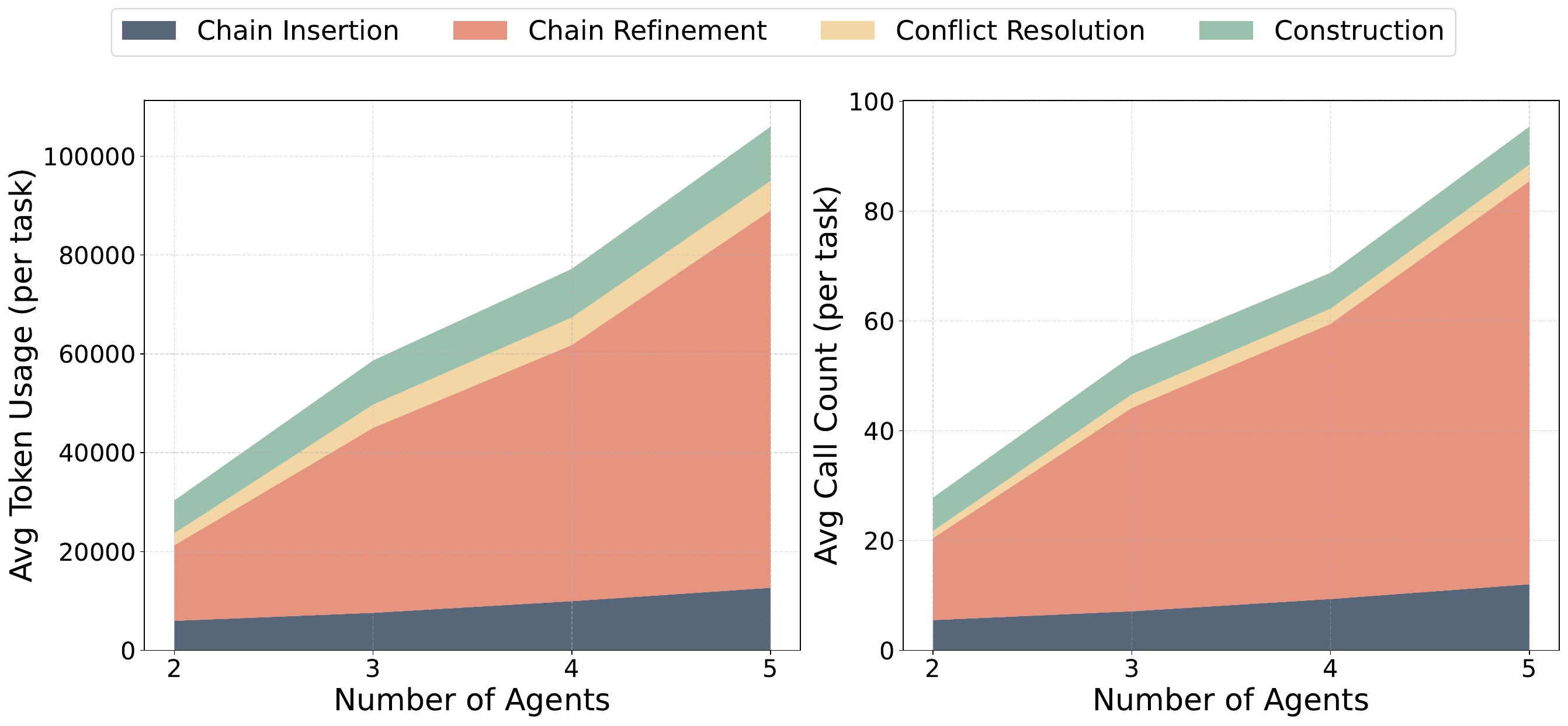}
    \caption{\hl{Average token usage (left) and LLM call count (right) per task broken down by pipeline stage, shown as a function of the number of agents. Chain Refinement accounts for the largest proportion across both metrics, while Chain Insertion contributes minimally.}} 
    \label{fig:stages_token_request}
\end{figure}

To further understand ELHPlan's resource allocation, Figure{~\ref{fig:stages_token_request}} breaks down token consumption and LLM call frequency across the four stages in multi-agent scenarios (2–5 robots). Chain Refinement dominates resource usage, accounting for 62.99\% of total tokens with 33.00 calls per \hl{task}, as it iteratively adjusts action chains throughout execution. In contrast, Construction and Chain Insertion consume moderate resources, Conflict Resolution accounts for merely 7.25\% with only 2.03 calls per \hl{task}. The low conflict resolution overhead confirms that the proactive action chain mechanism effectively prevents inter-agent conflicts, reducing the need for costly reactive re-planning.









\subsection{Ablation Study}
\label{sec:ablation_study}
\begin{figure}
    \centering
    \includegraphics[width=0.5\textwidth]{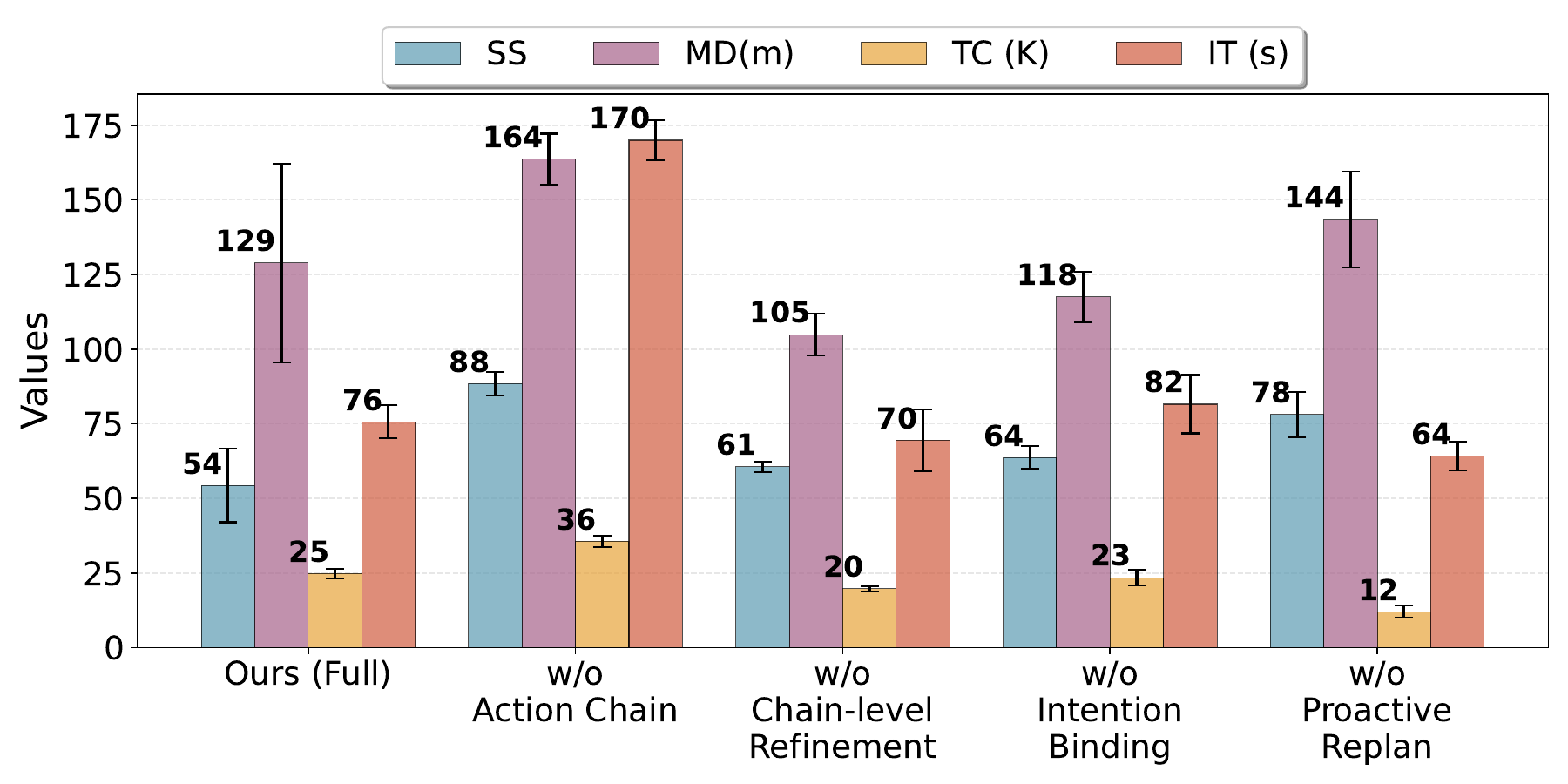}
    \caption{Ablation study results on the C-WAH benchmark with 2 agents using gpt-4o as the back bonne. }
    \label{fig:ablation_ex}
\end{figure}

To better understand the contribution of Action Chain and the refinement module in our framework, we conduct a series of detailed ablation experiments as shown in Figure~\ref{fig:ablation_ex}. Specifically, we disable components of Action Chain and refinement module and evaluate the performance on the multi-agent long-horizon task benchmark. 

We compare the following settings:
\begin{itemize}
    \item \textbf{Ours (Full)}: ELHPlan with Action Chain and all modules enabled. 
    \item \textbf{w/o Action Chain}: Eliminates the sequential action planning mechanism, forcing agents to plan step-by-step without chain-level coordination. 
    \item \textbf{w/o Chain-level Refinement}: Restricts the refinement process to single-action modifications, preventing holistic optimization of entire Action Chains.
    \item \textbf{w/o Intention Binding}: Generates Action Chains without explicit sub-goal intentions, removing the semantic coherence that guides refinement decisions.
    \item \textbf{w/o Proactive Replan}: Disables the anticipatory replanning mechanism, limiting replanning to reactive responses after execution failures or inter-agent conflicts.
\end{itemize}


\textbf{Action Chain Critical Impact.}  
Removing Action Chains causes the most severe degradation among all ablated components, with SS increasing by 62.9\% and IT nearly doubling (+123.7\%). 
Without multi-step commitment, agents revert to reactive single-step planning that requires a full LLM query when action finished, compounding both token consumption and coordination overhead. 

\textbf{Chain-level Refinement Value.} 
Restricting refinement to single-action modifications increases SS by 13.0\%, indicating that holistic chain-level refinement contributes to more efficient task completion. 
Notably, this variant yields lower TC and MD than the full model, as it forgoes the additional LLM queries needed to re-evaluate entire action sequences. 
This trade-off suggests that chain-level refinement's primary value lies in investing extra tokens to enable adaptive recovery from suboptimal states.

\textbf{Intention Binding Necessity.} Without intention annotations, SS increases by 18.5\%. 
Intention binding provides semantic anchoring needed to distinguish complementary from conflicting sub-goals, without which agents default to  conservative plans that  fail to take cooperative opportunities. 

\textbf{Proactive Replanning Benefits.} Disabling proactive replanning increases SS by 44.4\% while TC decreases by 52.0\%. 
Proactive replanning deliberately invests additional tokens to anticipate failures before they manifest, yielding a disproportionately large reduction in simulation steps.

\subsection{Limitations} 

While our approach demonstrates promising results, two key limitations remain that open avenues for future research. 
Although our current method is faster than other LLM-based iterative planning methods, generating long-horizon action sequences still requires more time than human planning, limiting the system’s ability to respond with human-like immediacy. 
Future work could explore model compression techniques and efficient inference methods to improve scalability and resource efficiency in large-scale deployments. 
Additionally, we occasionally observed cases where the model generates actions that violate provided constraints or exhibit hallucination-like behaviors. 
Addressing this issue may require incorporating reinforcement learning with human or preference feedback, in order to improve reliability and adherence to task specifications.

\section{Conclusion}

In this work, we present ELHPlan, a token-efficient framework that leverages Action Chain for robot coordination and planning. 
We introduce Action Chain, a novel representation that binds action sequences with explicit sub-goal intentions. 
ELHPlan overcomes the seemingly fundamental trade-off between planning efficiency and environmental adaptability through principled abstraction at the intention level, opening up new possibilities for practical large-scale multi-agent deployments. 
Future work will focus on training and fine-tuning lightweight models specifically optimized for multi-agent task planning using Action Chain. This will further improve planning speed and efficiency, while enabling scalability to large-scale multi-agent teams and practical deployment on physical robotic systems.

\bibliographystyle{IEEEtran}
\bibliography{reference.bib}

\appendix

\section*{Prompt Templates}

This supplementary material provides the complete prompt templates used in our framework. 
All prompts are reported verbatim to ensure reproducibility and transparency.




\begin{promptbox}[label={prompt:planning}]{Prompt Template for Task Assignment}
\begin{lstlisting}[style=promptstyle]
You are provided with a detailed task assignment.
<Examples> <Agent0 State <Agent1 State> <Agent1 Action Space> <Global Goal>
Please allocate tasks to agents.
...
Please think step by step. Then, generate the task assignment in the following JSON format:
{
  "reason": "<specific reason explanation>",
  "robot_id_task_pairs": [
    {
      "robot_id": "<robot_id>",
      "intention": "The intention or reason of the action chain.", 
      "action_chain": [
        "action from action space",
        "action from action space"
      ]
    },...
  ]
}
\end{lstlisting}

\end{promptbox}



\begin{promptbox}[label={lst:prompt_chain_refinement}]{Prompt Template for Chain Refinement}
\begin{lstlisting}[style=promptstyle]
Since the latest unexecuted action failed, it is best to realign the current action chain according to the original intention by refining it using the actions available in the action space.

Put the action `replan' right next to the agent who enters unexplored rooms.
<Examples> <Action Space> <Goal> <Observation> <Action Chain>
Ensure:
All rooms/objects use explicit names + IDs.
Actions should be selected from the action space.
Please think step by step.
Then, generate the output in the following JSON format:
{
    "inference_process": "The inference process of the better action chain.",
    "intention": "The intention of the action chain.",
    "action_chain": ["action", "action"]
}
\end{lstlisting}
\end{promptbox}



\begin{promptbox}[label={lst:prompt_chain_insertion}]{Prompt Template for Chain Insertion}
\begin{lstlisting}[style=promptstyle]
You need to make a plan using the actions in the action space to interact with task-related objects in the room where the agent is located, and work towards achieving our goals.
<Goal> <Observation> <Action Space> <Executed Plans> <Action Chain>
Rules: 1. The agent can only hold two objects. 2. Provides room names and IDs in action to eliminate ambiguity. 3. Summarize the intention of the plan. 4. Try to plan actions that are related to the current room.
Please think step by step. 
Then, generate the action chain for the agent to execute in the following JSON format:
{
     "inference_process": "The inference process of the action chain.",
     "intention": "The intention or reason of the action chain.",
      "action_chain": ["action", "action"]
}
\end{lstlisting}

\end{promptbox}


\begin{promptbox}[label={lst:prompt_conflict_resolution}]{Prompt Template for Chain Resolution}
\begin{lstlisting}[style=promptstyle]
Agent 0 and Agent 1 have conflicting action chains, meaning their historical and future actions overlap, resulting in unnecessary repetitions, such as repeatedly checking the same container or trying to grasp the same object.
Each agent can hold up to two objects in hand. 
Please evaluate the cost of each action in the thought process before moving 
on to planning.
<Examples> <Conflicting Action> <Agent0 Observation> <Agent0 Action Chain> <Agent1 Observation> <Agent1 Action Chain>
Determine the first planned action based on your current state. 
Then, consider the state after executing the action and conceive the next action. 
Think step by step.
Please generate the output in the following JSON format:
{
    "reason_agent0": "The reason for agent0 choosing the method. ",
    "action_chain0": ["action","action"],
    "intention0": "The intention of the action chain.",
    "reason_agent1" : "The reason for agent1 choosing the method. ",
    "action_chain1": ["action","action"],
    "intention1": "The intention of the action chain.",
}
\end{lstlisting}

\end{promptbox}

\end{document}